\def\year{2019}\relax
\documentclass[letterpaper]{article} 
\usepackage{aaai19}  
\usepackage{times}  
\usepackage{helvet}  
\usepackage{courier}  
\usepackage{url}  
\usepackage{graphicx}  
\usepackage{subfigure}
\usepackage{framed,multirow}
\usepackage{amssymb}
\usepackage{amsmath}
\usepackage{latexsym}
\usepackage{makecell}
\title{EnsNet: Ensconce Text in the Wild}
\author{Shuaitao Zhang$^*$, Yuliang Liu\thanks{Authors contributed equally as first author}, Lianwen Jin\thanks{Corresponding auther}, Yaoxiong Huang, Songxuan Lai\\
School of Electronic and Information Engineering\\
South China University of Technology\\
lianwen.jin@gmail.com\\
}
 
\begin{document}
\maketitle
\begin{abstract}
A new method is proposed for removing text from natural images. The challenge is to first accurately localize text on the stroke-level and then replace it with a visually plausible background. Unlike previous methods that require image patches to erase scene text, our method, namely ensconce network (EnsNet), can operate end-to-end on a single image without any prior knowledge. The overall structure is an end-to-end trainable FCN-ResNet-18 network with a conditional generative adversarial network (cGAN). The feature of the former is first enhanced by a novel lateral connection structure and then refined by four carefully designed losses: multiscale regression loss and content loss, which capture the global discrepancy of different level features; texture loss and total variation loss, which primarily target filling the text region and preserving the reality of the background. The latter is a novel local-sensitive GAN, which attentively assesses the local consistency of the text erased regions. Both qualitative and quantitative sensitivity experiments on synthetic images and the ICDAR 2013 dataset demonstrate that each component of the EnsNet is essential to achieve a good performance. Moreover, our EnsNet can significantly outperform previous state-of-the-art methods in terms of all metrics. In addition, a qualitative experiment conducted on the SMBNet dataset further demonstrates that the proposed method can also preform well on general object (such as pedestrians) removal tasks. EnsNet is extremely fast, which can preform at 333 fps on an i5-8600 CPU device.
\end{abstract}
\section{1. Introduction}
\label{sec:intro}
Scene text is ubiquitous in our daily life, and it conveys valuable information. However, various private information, such as ID numbers, telephone numbers, car numbers, and home addresses \cite{Inai2014Selective} may easily be exposed in natural scene images. Such important private information can be easily collected automatically by the machines engaged in fraud, marketing, or other illegal activities. Therefore, a method that can ensconce the text in the wild would be beneficial. In addition to preventing privacy disclosure, Scene text erasing can also facilitate many image processing and computer vision applications, such as information reconstruction and visual translation.
\begin{figure}[t]
\includegraphics[width=\linewidth]{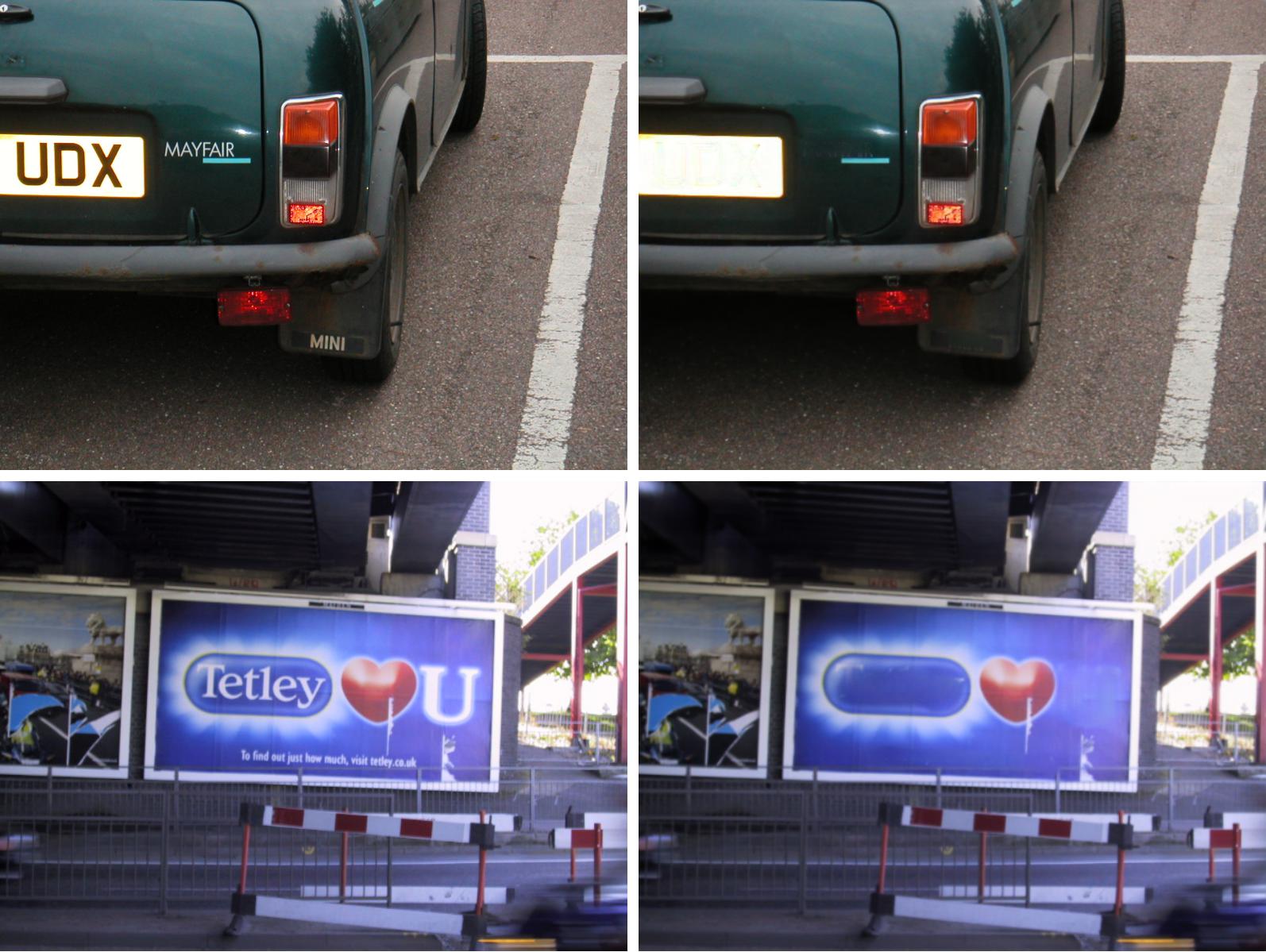}
\caption{Examples of scene text erasing. Given a scene text image, the goal is to ensconce the text and replace it with a visually plausible background while retaining the non-text regions. Left: input images. Right: our results.}
\label{fig:background}
\end{figure}

Examples of scene text erasing are shown in Fig~\ref{fig:background}. The challenges of erasing the scene text are as follows: 1) scene text erasing methods should be able to perceive the stroke-level position of the text in advance, which is more complicated than bounding-box-level scene text detection methods that have been comprehensively researched \cite{liang2005camera}; 2) after removing text, the original text region should be filled with a new visually plausible background; 3) the non-text regions should be retained in their original appearance.

To the best of our knowledge, the method proposed in \cite{Nakamura2017Scene} is the first and the only method that addresses the scene text erasing issue. However, their method uses cropped image patches as training data, which limits the erasing performance due to loss of global context information.

We herein propose a novel method named the ensconce network (EnsNet) to address this task, which includes the following characteristics:
\begin{itemize}
\item EnsNet can erase the scene text end-to-end on the whole image, which improves the erasing results and executes very fast.
\item A novel lateral connection structure is designed to effectively capture the detailed context information.
\item A refined loss function including multiscale regression loss, content loss, texture loss, and total variation loss (tv) loss is proposed in the optimization stage to ensure the background reconstruction and integrity of the non-text region.
\item A local-aware discriminator is proposed to guide the network to replace the text region with a more plausible background.
\end{itemize}

In addition, we also synthesized a dataset to benchmark the scene text erasing and facilitate future research.

The experimental results on both the synthetic dataset and the ICDAR 2013 dataset \cite{Karatzas2013ICDAR} demonstrate that the EnsNet can significantly outperform previous state-of-the-art methods in terms of all metrics.

\section{2. RELATED WORK}
\label{sec:related work}
In recent years, several methods have been proposed to remove graphic text from born-digital images such as captions, subtitles, and annotations \cite{khodadadi2012text,Modha2014Image,wagh2015text}. However, these methods required the text to be axis-aligned, clean, and well focused, therefore they are not applicable to scene text erasing because of its complexities, uneven illuminations, perspective distortions, etc.
\begin{figure*}[t]
\centering
\includegraphics[width=\textwidth]{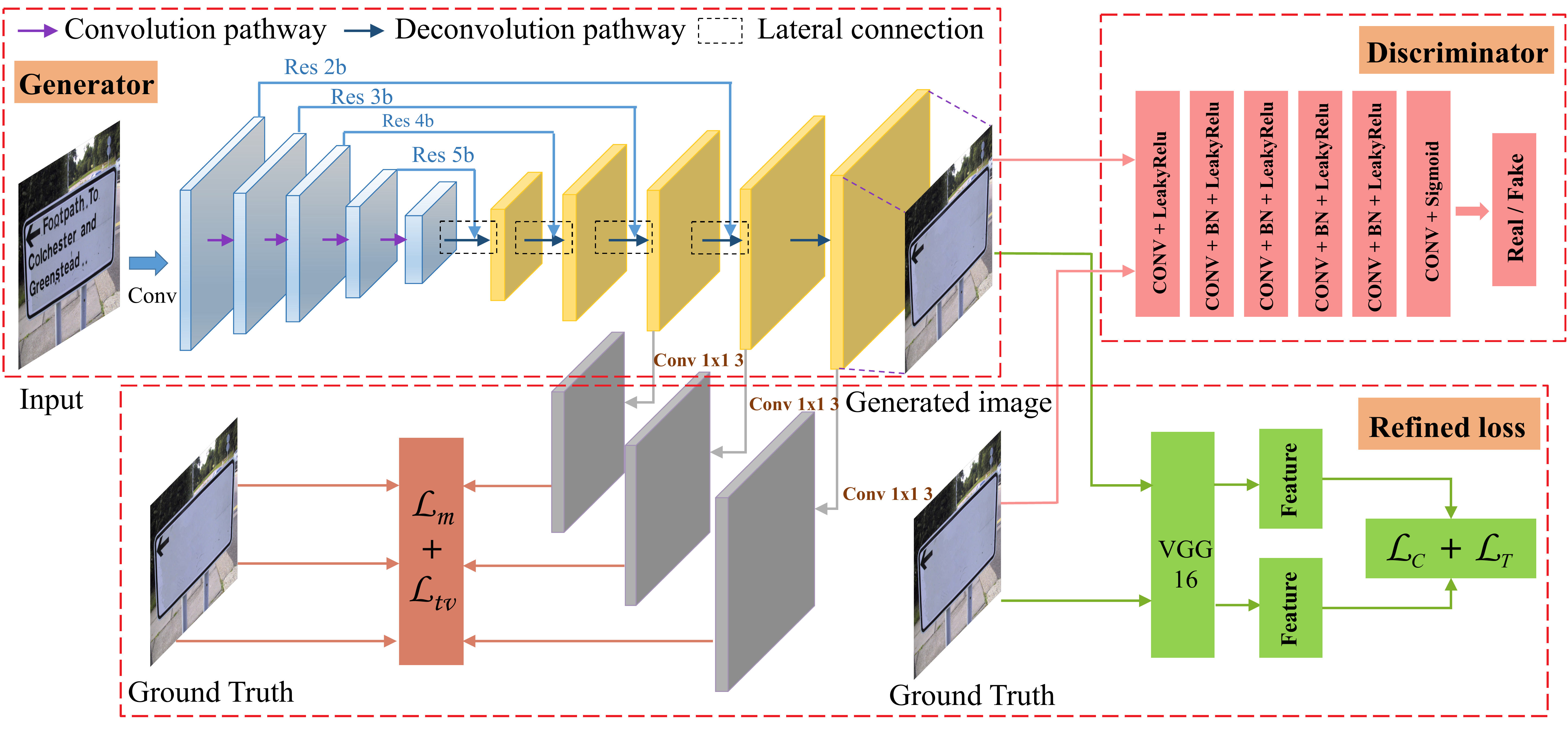}
\caption{The overall structure of EnsNet. The network consists of three seamlessly-connected components: 1) the refined generator backbone (enhanced by the lateral connections); 2) refined-loss modules; 3) EnsNet discriminator. The network is end-to-end trainable with extremely high efficiency.}
\label{fig:network}
\end{figure*}

Recently, owing to the immense success of deep learning in computer-vision tasks, \cite{Nakamura2017Scene} were the first to successfully design a Scene text eraser to ensconce the text in the wild. Specifically, this method used a single-scaled sliding-window-based neural network to erase the scene text. However, Nakamura et al.'s method is prone to dividing large text into multiple image patches, thus significantly reducing the consistency of the erasing results. In addition, using cropped image patches as training data will restrict the operating speed to some extent.

\cite{isola2017image} proposed an extremely fast Pix2Pix method that not only learns the mapping from the input image to output image, but also introduces a novel loss function to train the mapping. Although this state-of-the-art method is not proposed specifically to handle the scene text erasing task, it can be easily reproduced for comparison.

\section{3. METHODOLOGY}
\label{sec:methodology}
\subsection{3.1. Framework Overview}
Fig. \ref{fig:network} shows the overall architecture of the proposed EnsNet, which consists of two primary parts: a generator G and a discriminator D. Given a scene text image \textbf{x} and the ground truth \textbf{z}, EnsNet attempts to produce a non-text image \textbf{y} that is as real as possible by solving the following optimization problem:
\begin{equation}
\begin{split}
   \min\limits_{G} \max\limits_{D} \mathbb{E}_{\textbf{x}\sim p_{data(\textbf{x})},\textbf{z}}\left[\log(1-D(\textbf{x},G(\textbf{x},\textbf{z}))) \right] +\\
\mathbb{E}_{\textbf{x}\sim p_{data(\textbf{x},\textbf{y})},\textbf{z}}\left[\log D(\textbf{x},\textbf{y})\right].
\end{split}
\end{equation}

Following the ideas of previous GAN-based methods \cite{Mirza2014Conditional,Goodfellow2014Generative}, the proposed method alternatively updates G and D, and the entire procedure is end-to-end trainable.

\subsection{3.2. Generator}
The construction of G contains three mutually promoted modules: a lightweight Fully-Convolution-Network (FCN) backbone, lateral connections, and refined loss.

\subsubsection{Fully convolution network}
The FCN consists of a convolution-pathway and a deconvolution-pathway. The former utilizes a lightweight Resnet18 network. Based on the last convolutional layer of the Resnet18, a 1$\times$1 convolutional layer converted by the last fully-connected layer is applied to predict the text/non-text score map. The deconvolution pathway consists of five deconvolutional layers, with kernel size set 4, stride step set 2, and padding size set 1 for each layer.
\begin{figure}[h]
\centering
\includegraphics[width=0.45\textwidth]{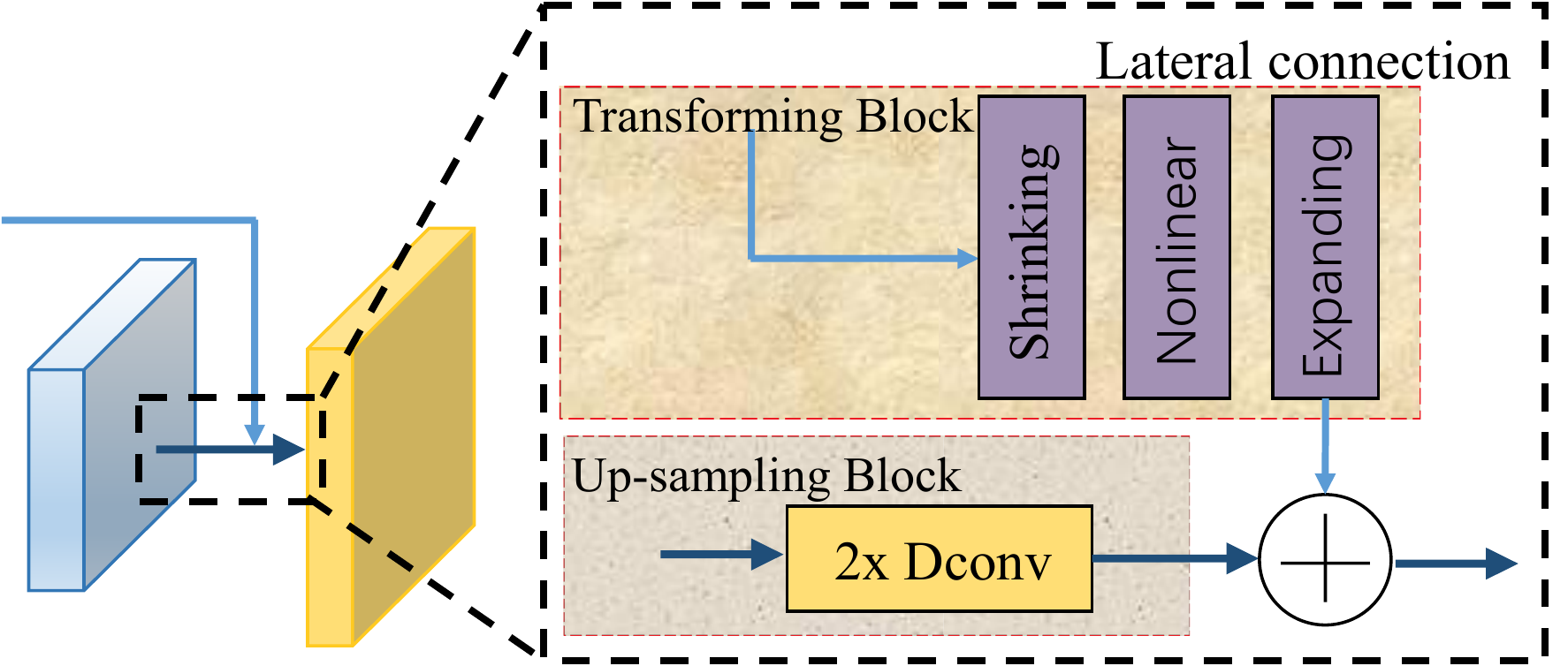}
\caption{The structure of lateral connection.}
\label{fig:lateralloss}
\end{figure}

\subsubsection{Lateral connections}
It is typically considered that lower level features exhibit stronger semantics, while higher level exhibit semantically weak features but more detailed information including pixel-level color, texture, the position information of objects, etc \cite{Shen2017Convolutional,dong2016accelerating}. Hence, we designed the lateral connections, as shown in Fig. \ref{fig:lateralloss}, to integrate the higher-level semantics with details from the lower layers. The proposed lateral connection includes a transforming block and an up-sampling block:

1) The transforming block starts with a shrinking layer, that reduces the feature dimensions by using a 1$\times$1 convolution. Subsequently, two same-size (3$\times$3) convolutional layers are stacked to perform a nonlinear transformation, that can not only replace large-kernel convolutions \cite{Dong2016Image} to achieve large receptive fields, but also improve the computation efficiency. Finally, we use an expanding layer to enlarge the feature map channels by a 1$\times$1 convolution as reverse of the shrinking operation. The transformation block takes the Residual2b to the Residual5b of the Resnet18 as input.

2) For the up-sampling block, we used a deconvolutional layer to enlarge the feature map. The up-sampled features maps are then element-wise summed with the corresponding ones from the transforming block.
Additionally, unlike previous CNN-based methods, we replaced all ReLU/LeakyReLU layers with the ELU layers \cite{clevert2016fast,Yang2016High} after each layer except the layers in the convolution pathway. The ELU layers render the generator network training more stable compared with the ReLU/LeakyReLU layers \cite{Nair2010Rectified} as it can handle large negative responses during the training process.

\subsubsection{Refined loss}
Our loss functions target both per-pixel reconstructed accuracy as well as composition, i.e. how smoothly the text regions can harmonize with their surrounding context. As shown in Fig. 2, there are four functions designed for our generative network: multiscale regression loss, content loss, texture loss, and total variation loss.

{\bf{1) The multiscale regression loss.}} Given an input image with text $\mathcal{I}_{in}$, initial binary mask $\mathcal{M}$ (0 for non-text regions and 1 for text regions), the generator prediction $\mathcal{I}_{out}$, and the ground truth image $\mathcal{I}_{gt}$, we extracted features from different deconvolutional layers to form outputs of different sizes. By adopting this, more contextual information from different scales can be captured.

The multiscale regression loss function is defined as:
\begin{equation}
\begin{split}
    \mathcal{L}_m \left( \mathcal{M}, \mathcal{I}_{out},\mathcal{I}_{gt} \right)=\sum_{i=1}^{n} \lambda_i (\lVert\mathcal{M}_i\odot(\mathcal{I}_{out(i)}-\\ \mathcal{I}_{gt(i)})\lVert_1
 + \alpha \lVert(1-\mathcal{M}_i) \odot(\mathcal{I}_{out(i)}- \mathcal{I}_{gt(i)}) \lVert_1 ),
\end{split}
\end{equation}
where $\mathcal{I}_{out(i)}$ indicates the $ith$ output extracted from the deconvolutional pathway, $\mathcal{M}_i$, $\mathcal{I}_{gt(i)}$ indicates the binary mask and ground truth that have the same scale as that of $\mathcal{I}_{out(i)}$ respectively, and $\alpha$ weights the importance between text and non-text regions. $\lambda_i$ is the weight for the $ith$ scale. Concretely, the output of the last, and $3^{rd}$ and $5^{th}$ last layers are used, whose sizes are $\frac{1}{4}$, $\frac{1}{2}$, and 1 of the input size, respectively. Practically, we set $\alpha$ to 6 and $\lambda_i$ to 0.6, 0.8, 1.0 to put more weight at the larger scale. L1 loss is adopted on the network output for the text regions and the non-text regions, respectively.

{\bf{2) The content loss.}} Recently, a loss function measured on
different high-level features has been demonstrated effective for feature reconstructing \cite{Johnson2016Perceptual}. To further enhance the text erasing performance, we introduce content constraints on high-level features that we term as content loss. The content loss penalizes the discrepancy between the features of the output image and the corresponding ground truth image on certain layers in the CNN. We feed an output image and the corresponding ground truth to the CNN, and enforce the response of the output image to match that of the ground truth at the predetermined feature layers of the CNN, which will facilitate the network in detecting and erasing text regions. The content loss is defined as follows:
\begin{equation}
\begin{split}
    \mathcal{L}_{C} = \sum_{n=1}^{N-1}\lVert(\mathcal{A}_n(\mathcal{I}_{out})-\mathcal{A}_n(\mathcal{I}_{gt})\lVert_1 \\
 + \sum_{n=1}^{N-1}\lVert(\mathcal{A}_n(\mathcal{I}_{comp})-\mathcal{A}_n(\mathcal{I}_{gt})\lVert_1.
\end{split}
\end{equation}
Where, $\mathcal{I}_{comp}$ is the output image $\mathcal{I}_{out}$, with the non-text regions of $\mathcal{I}_{out}$ being set to the ground truth. $\mathcal{A}_n$ is the activation map of the n-th selected layer. In our method, we compute the feature loss at layers pool1, pool2, and pool3 of a pretrained VGG16 \cite{Simonyan2014Very}.

{\bf{3) The texture loss.}} As discussed earlier, the visual quality should also be considered into the optimization function. Hence, we introduce the texture loss that ensures that the restored text regions match with the non-text regions. The loss is motivated by the recent success of neural style transfers \cite{Gatys2015A}. Texture loss performs an autocorrelation (Gram matrix) \cite{Gatys2016Image} on each high-level feature map before applying the L1 loss, which can be defined as follows:
\begin{equation}
\begin{split}
    \mathcal{L}_{T_{out}} = \sum_{n=1}^{N-1}\lVert\frac{1}{C_nH_nW_n}((\mathcal{A}_n(\mathcal{I}_{out}))^T\\(\mathcal{A}_n(\mathcal{I}_{out}))
 - ((\mathcal{A}_n(\mathcal{I}_{gt}))^T(\mathcal{A}_n(\mathcal{I}_{gt}))\lVert_1,
\end{split}
\end{equation}
\begin{equation}
\begin{split}
    \mathcal{L}_{T_{comp}} = \sum_{n=1}^{N-1}\lVert\frac{1}{C_nH_nW_n}((\mathcal{A}_n(\mathcal{I}_{comp}))^T\\(\mathcal{A}_n(\mathcal{I}_{comp}))
 - ((\mathcal{A}_n(\mathcal{I}_{gt}))^T(\mathcal{A}_n(\mathcal{I}_{gt}))\lVert_1,
\end{split}
\end{equation}
where the (H$_n$W$_n$) $\times$ C$_n$ is the shape of the high-level activation map, $\mathcal{A}_n$. Similar to the idea proposed in \cite{Zhang2017Image} , we aim to penalize the discrepancy of the texture appearance of text regions and non-text regions, such that the network can capture the globally style features for more reasonably displacing the text regions. Again, we include loss terms for the raw output $\mathcal{I}_{out}$ and the output of text-erased regions $\mathcal{I}_{comp}$.

{\bf{4) The total variation loss.}} The last $\mathcal{L}_{tv}$ \cite{Johnson2016Perceptual} is targeted at global denoising, as defined below:
\begin{equation}
\begin{split}
    \mathcal{L}_{tv} = \sum_{i,j}^{}\lVert\mathcal{I}_{out}^{i,j}-\mathcal{I}_{out}^{i+1,j}\lVert_1+ \lVert\mathcal{I}_{out}^{i,j}-\mathcal{I}_{out}^{i,j+1}\lVert_1,
\end{split}
\end{equation}
where i, j is the pixel position.

\begin{figure}[htb]
\includegraphics[width=\linewidth]{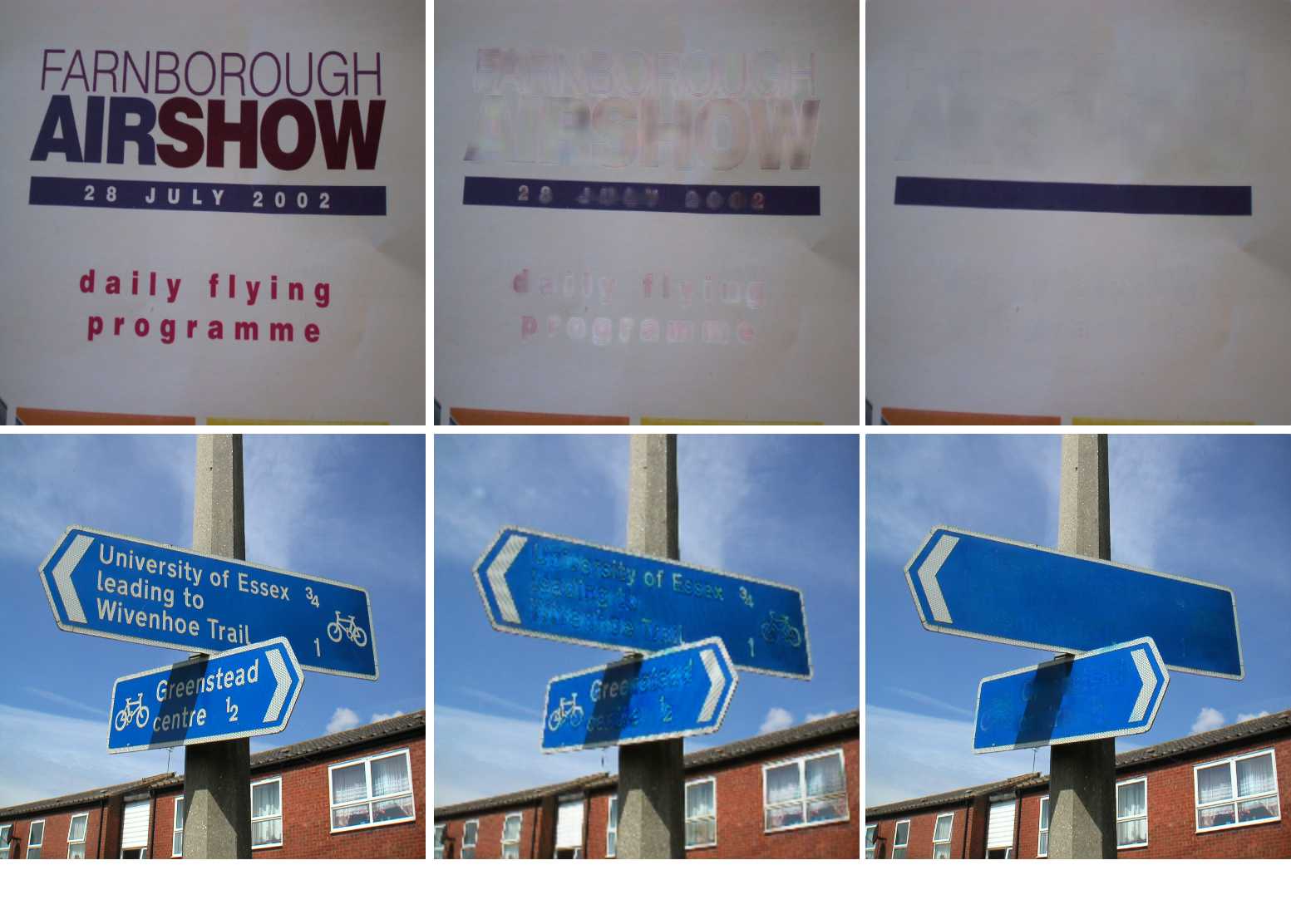}
\caption{In the top row, images from left to right: input image; result with small content loss and texture loss weights; result with refined loss. In the bottom row, images from left to right: input image; result using large content loss and texture loss weights; result with refined loss.  }
\label{fig:resLoss}
\end{figure}

To exploit all merits, we combine these four novel losses together with the appropriate weights to form the refined loss function, which is defined as the following:
\begin{equation}
\label{eq:refineLoss}
\begin{split}
    \mathcal{L}_{refined} = \mathcal{L}_{M}+\lambda_e\mathcal{L}_{C}+\lambda_i\mathcal{L}_{T}+\lambda_t\mathcal{L}_{tv}.
\end{split}
\end{equation}
The hyper-parameters $\lambda_e$, $\lambda_i$, and $\lambda_t$ in Eq. \ref{eq:refineLoss} control the balance between the four losses. For our model, not all the loss weighting schemes for the refined loss will generate satisfactory results. Fig. \ref{fig:resLoss} shows the result of the model trained with small or large weights of content loss and texture loss. In our experiments, $\lambda_e$, $\lambda_i$, and $\lambda_t$ are empirically set to 0.5, 50.0, and 25.0, respectively. It is noteworthy that such parameters setting works well when generalizing to other tasks, which is discussed in Section. 4.5.
\subsection{3.3. Discriminator}
To differentiate fake images from real ones, the original GANs \cite{Goodfellow2014Generative} discriminate the results based on the whole image level. However, the non-text regions occupy a large proportion of the image and are typically real, which makes the discriminator difficult to focus on the text regions. Therefore, it is intuitive to design a local-aware discriminator to attentively maintain the consistency of text-erased regions and their surrounding texture.

The proposed local-aware discriminator only penalizes the erased text patches. Specifically, following the PatchGAN \cite{isola2017image}, we execute our discriminator across an image and obtain an S$\times$S (S=62) feature tensor. By designation, each entry of the tensor corresponds to a N$\times$N (N = 70) patch (i.e., receptive filed) of the image, and is assigned a label to indicate its realness. The key novelty of our discriminator is the labeling strategy. For the typical condition discriminative network \cite{isola2017image}, they will assign negative labels for any generated images, which is contrary to our purpose. Therefore, we assign a locality-dependent label for the each position of the 62*62 tensor according to the text mask.
\begin{equation}
\begin{split}
label =
\begin{cases}
0,  & \mbox{if }sum (\mathcal{M})\mbox{ $>$ 0,} \\
1, & \mbox{otherwise.}
\end{cases}
\end{split}
\end{equation}
Where $\mathcal{M}$ is the binary mask (as mentioned in the above section). Given the discriminator prediction $\mathcal{P}_i$, and the label $\mathcal{L}_i$ for each N $\times$ N patch, we define the loss as follows:
\begin{equation}
\begin{split}
\mathcal{L}_{D} =
-\sum_{i=1}^{S^2}\frac{sum(\mathcal{M}_i)}{N \times N}(1-\mathcal{L}_i)(\log(\mathcal{P}_i)).
\end{split}
\end{equation}
The scaling factor sum ($\mathcal{M}_i$) applies the appropriate scaling to adjust for the varying amount of text-erased regions. Our underlying idea is to guide our discriminator to focus on regions indicated by the text mask. The structure of the discriminative architecture is shown in Fig. \ref{fig:network}.
\begin{figure*}[t]
\centering
\includegraphics[width=\textwidth]{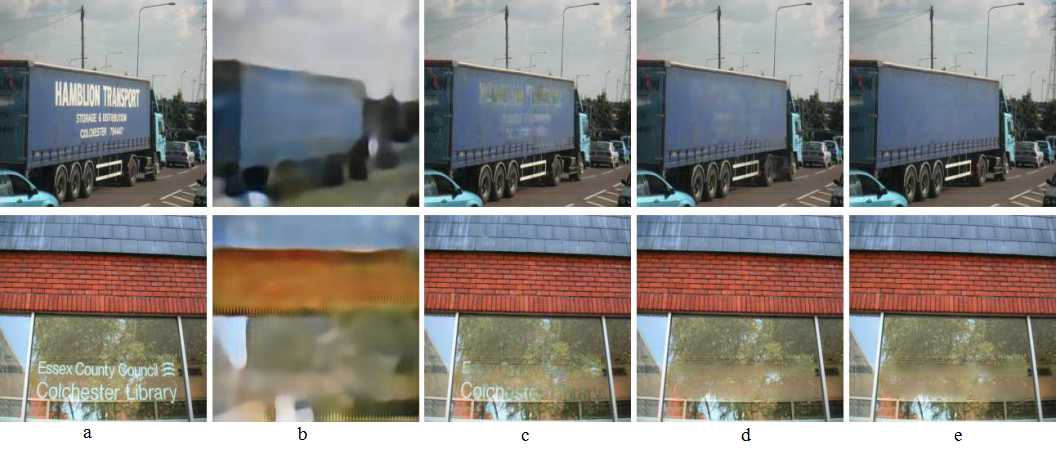}
\caption{Illustration of improvements obtained using different components of our network. Images from left to right: (a): input images; (b): erased results by baseline; (c): erased results by lateral connections; (d): erased results of refined loss; (e): erased result by the complete architecture (EnsNet).
}
\label{Fig:resComponents}
\end{figure*}
\section{4. EXPERIMENT}
To quantitatively evaluate the performance of scene text erasing methods, a dataset should provide both text images with stroke-level ground truths and real background images without text. However, such text dataset does not exist currently; thus, we constructed a synthetic dataset for evaluating the performance. In addition, we evaluated the performance on the ICDAR 2013 \cite{Karatzas2013ICDAR} dataset following the same metric as reported by ~\cite{Nakamura2017Scene} for a fair comparison; the metric utilizes an auxiliary detector to validate whether the text of images can be detected after processing by the erasing method. An ablation study is also conducted to evaluate each component of EnsNet.

\subsection{4.1. Dataset and evaluation metrics}
\subsubsection{Synthetic data}
\begin{figure}[htb]
\begin{minipage}[b]{\linewidth}
  \centering
  \centerline{\includegraphics[width=8.0cm,,height=4.0cm]{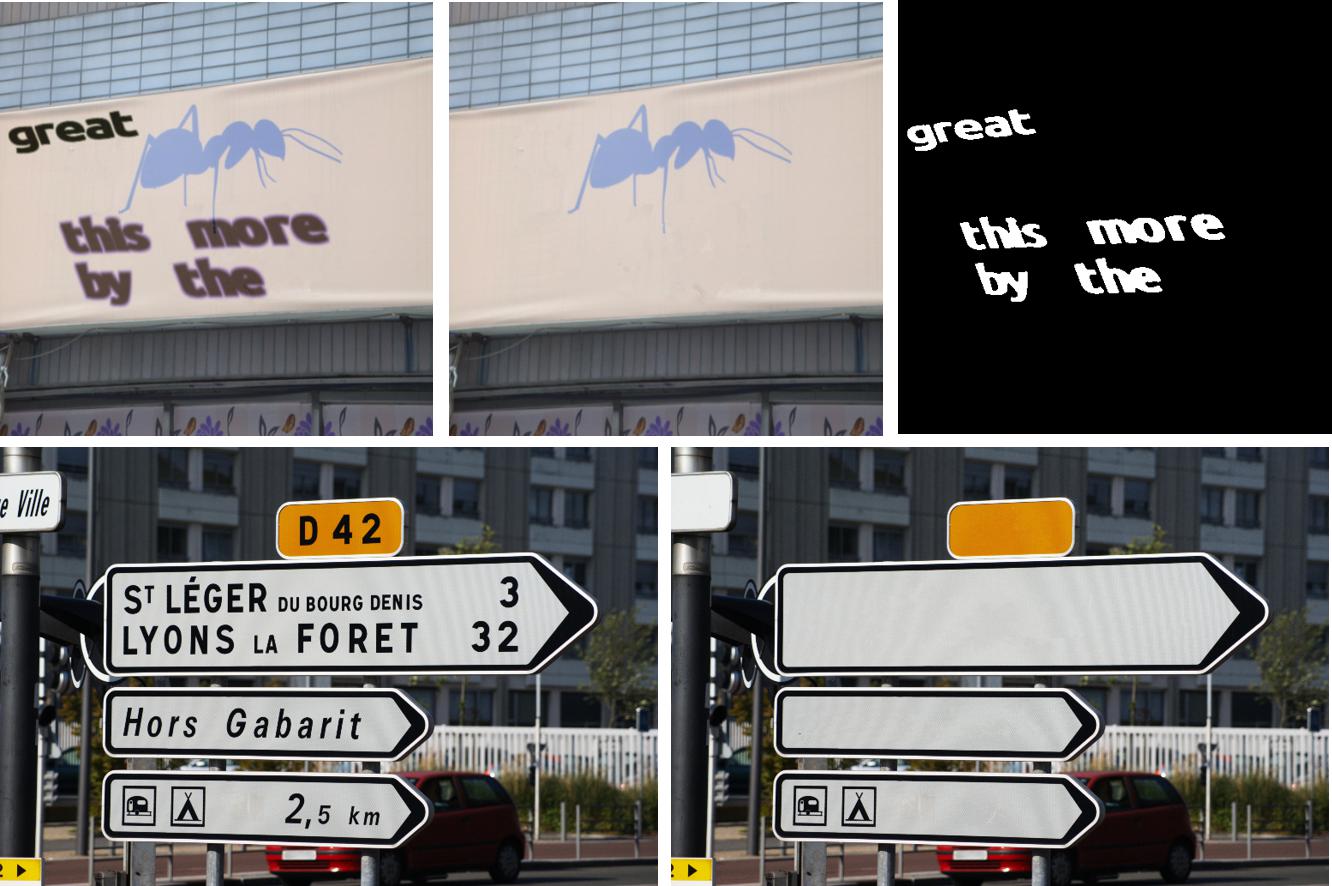}}
\end{minipage}
\caption{Examples of training samples for EnsNet learning. Top: synthetic sample. Bottom: real sample.}
\label{fig:datasetGeneration}
\end{figure}

We apply text synthesis technology \cite{Gupta2016Synthetic} on scene images to generate samples, as shown in the first row of Fig.~\ref{fig:datasetGeneration}. Compared to the inpainting algorithm \cite{Criminisi2004Region}, the label generated by the synthesis process is more accurate and reasonable. In our experiments, the training set consists of a total of 8000 images and the test set contains 800 images; all the training and test samples are resized to 512 $\times$ 512. The synthetic data are available at: https://github.com/HCIILAB/Scene-Text-Removal.

\subsubsection{Real-word dataset} ICDAR 2013 \cite{Karatzas2013ICDAR} is a widely used dataset for scene text detection, containing 229 images for training, and 223 for testing. ICDAR 2017 MLT \cite{nayef2017icdar2017} is a benchmark multi-lingual scene text dataset. In our experiment, we collected 1000 images from the ICDAR 2017 MLT subdataset which only contains English text to enlarge real data, and the background image (label) is generated by manually erasing the text. An example is shown in Fig. \ref{fig:datasetGeneration}. For a fair comparison, all our method and counterpart methods strictly use the same training set and hyper-parameters.

\subsubsection{Evaluation metrics}
\cite{Nakamura2017Scene} proposed a new evaluation method that utilizes a baseline detector to adversely evaluate the erased results by computing how low the precision is, as well as the recall and f-score of the ICDAR 2013 test set \cite{Karatzas2013ICDAR}. However, this method does not consider the quality of the erased image, i.e, if the image contents are completely erased, the result would be the best, which is not reasonable. Therefore, to impartially compare the proposed method with other methods on the synthetic dataset, we adopted the previous image inpainting \cite{Yang2016High,liu2018image} metrics to evaluate our method, which includes the following: 1) $\ell_2$ error, PSNR, which computes the peak signal to noise ratio (PSNR) for an image; 2) SSIM \cite{Wang2004Image}, which computes the mean structural similarity index between two images; 3) AGE, which calculates the average of the gray-level absolute difference between the ground truth and the computed background image; 4) pEPs, which calculate the percentage of error pixels; 5) pCEPS, which calculates the percentage of clustered error pixels (number of pixels whose four-connected neighbors are also error pixels). A higher SSIM and PSNR or lower AGE, pEPs, pCEPS and the $\ell_1$ error represent better results. For the real dataset, because we do not have ground truth background images, we follow the same procedure as reported by \cite{Nakamura2017Scene} to calculate the precision, recall, and f-score. Additionally, visual estimates are also used on real dataset to qualitatively compare the performance of various methods.
\subsection{4.2. Ablation study}
In this section, we apply the sensitivity analysis to verify the contributions of different components of EnsNet:
{\bf {Baseline}}, {\bf {Lateral Connections}}, {\bf {Refined Loss}}, and {\bf {EnsNet}}.

All experiments use exactly the same settings (input size is set to 512 $\times$ 512). The quantitative results are shown in Tab. \ref{tab:components}, and the qualitative results can be visualized in Fig. \ref{Fig:resComponents}.
\subsubsection{Baseline}
We directly apply fully convolution network to train baseline model only with L1 loss.
\subsubsection{Lateral connections}
In Tab. \ref{tab:components}, compared with the baseline model, the lateral connections obtain a lower $\ell_2$ error. This implies that the lateral connections can effectively learn more detailed information to restore the image detail. Meanwhile, from the Fig. \ref{Fig:resComponents}, we found that lateral connections are conducive for our generative network to achieve clearer results.
\subsubsection{Refined Loss}
Tab. \ref{tab:components} shows that the refined loss can improve erasing result on any type of metric mentioned above. Some intuitive examples are shown in Fig. \ref{Fig:resComponents}, where the fish scale artifacts or blocky checkerboard artifacts caused by cGAN are ameliorated by our refined loss, thereby demonstrating the effectiveness of using the proposed refined loss.
\subsubsection{EnsNet}
By seamlessly connecting with the proposed local-awareness discriminative network, the performance of EnsNet can be further improved, as shown in both Tab. \ref{tab:components} and Fig. \ref{Fig:resComponents}. EnsNet alleviates the dilemma of the extreme imbalance between the text regions and the non-text regions; thus, it can attentively replace the text with a more plausible background.
\begin{table*}[!htb]
\caption{Quantitative evaluation results of different components of our method and other methods on synthetic and real datasets. FCN: fully convolution network; LC: lateral connections; RL: refined loss; TD: text discriminator; R: recall; P: precision; F: f-measure. RPF metric is calculated by ICDAR 2013 Evaluation, and DSSD \cite{Fu2017DSSD} is adopted as an auxiliary detector. Note that although the results of the baseline contain lower values in RPF, it is significantly worse than other methods in terms of other metrics such PSNR and SSIM. This implies that the previously used RPF metric is not a perfect metric (visualization results and other metrics should also be taken into considered).}
\tiny
\label{tab:components}
\centering
\begin{tabular}{|c|>{\makecell*[c]}c|c|c|c|c|c|c|c|c|c|c|c|c|c|c|c|c|}
\hline
\multirow{2}{*}{Method} &\multicolumn{1}{c|}{\multirow{2}{*}{FCN}}&\multicolumn{1}{c|}{\multirow{2}{*}{LC}} &\multicolumn{1}{c|}{\multirow{2}{*}{RL}}&\multicolumn{1}{c|}{\multirow{2}{*}{TD}} &\multicolumn{9}{c|}{Synthesis} &\multicolumn{3}{c|}{ICDAR 2013}&\multicolumn{1}{c|}{\multirow{2}{*}{FPS}}\\
\cline{6-17}
\multicolumn{1}{|c|}{}&\multicolumn{1}{c|}{}&\multicolumn{1}{c|}{}&\multicolumn{1}{c|}{}&\multicolumn{1}{c|}{} &PSNR  &SSIM (\%)  &$\ell_2$ error  &AGE &pEPs &pCEPS  & R &P & F& R &P & F &\multicolumn{1}{c|}{} \\
\hline
\makecell[{tc}]{Original \\ images}  &-&-  &-  &-  & - & - & - & - & - & - & 53.65 & 68.21 & 60.06 & 86.11 & 95.66 & 90.64 &- \\
\hline
Baseline &$\surd$&  &  &  & 20.30 & 62.18 & 0.4256&13.84&0.1918&0.1091 & 0.10 & 1.41 & 0.19 &2.28  &41.67  & 4.33 &- \\
\makecell[{tc}]{Lateral \\ Connections} &$\surd$&$\surd$  &   &   & 24.83 & 86.74 & 0.0627&5.48&0.0538&0.0385 & 26.72 & 40.89 & 32.32   & 51.91 & 79.20 & 62.71&-\\
Refined Loss &$\surd$&$\surd$ &$\surd$ &  & 30.23 & 92.78 & 0.0234 &2.51 &0.0227 &0.0121 & 8.35 & 26.59 & 12.71  & 14.64 & 72.21 & 24.34 &-\\
EnsNet &$\surd$&$\surd$ &$\surd$ &$\surd$ & \textbf{37.36} & \textbf{96.44} & \textbf{0.0021} &\textbf{1.73}&\textbf{0.0069}&\textbf{0.0020}& \textbf{1.14} & \textbf{23.45} & \textbf{2.17} & \textbf{12.58} & \textbf{70.00} & \textbf{21.33} & 333 \\
\hline
\makecell[{tc}]{Erased by \\ Pix2Pix \shortcite{isola2017image}} & -& - & - & - & 25.60 & 89.86 & 0.2465 &5.60&0.0571&0.0423& 28.31 & 65.30 & 39.50 & 32.33 & 85.08 & 46.84 & \textbf{382}\\
\hline
\makecell[{tc}]{Erased by Scene\\ text eraser \shortcite{Nakamura2017Scene}} & -& - & - & - & 14.68 & 46.13 & 0.7148&13.29&0.1859&0.0936 &22.35 & 30.12 & 25.66 &34.48 & 60.57 & 43.95 & 166 \\
\hline
\end{tabular}
\end{table*}

\begin{table*}[!htb]
\caption{Comparison among previous methods and proposed method. R: Recall; P: Precision; F: F-measure.}
\scriptsize
\label{tab:compareSOTA}
\centering
\begin{tabular}{|c|c|c|c|c|c|c|c|c|c|c|c|c|c|}
\hline
\multirow{3}{*}{Detection} &\multicolumn{1}{c|}{Image dataset}&\multicolumn{6}{c|}{Synthesis}&\multicolumn{6}{c|}{ICDAR 2013}\\
\cline{2-14}& Evaluation protocol&\multicolumn{3}{c|}{ICDAR Eval}&\multicolumn{3}{c|}{DetEval}&\multicolumn{3}{c|}{ICDAR Eval}&\multicolumn{3}{c|}{DetEval}\\
\cline{2-14}&Methods&R&P&F&R&P&F&R&P&F&R&P&F \\
\hline
\multirow{4}{*}{DSSD \shortcite{Fu2017DSSD}} & Original image & 53.65 & 68.21 & 60.06 & 55.19 & 68.94 & 61.30 & 86.11 & 95.66 & 90.64 & 86.08 & 95.87 & 90.71 \\
\cline{2-14}
&Erased by Pix2Pix \shortcite{isola2017image} & 28.31 & 65.30 & 39.50 & 29.54 & 65.89 & 40.79 & 32.33 & 85.08 & 46.85 & 33.97 & 85.95 & 48.70 \\
&Erased by Scene text eraser \shortcite{Nakamura2017Scene}& 22.35 & 30.12 & 25.66 & 22.67 & 30.76 & 26.10 & 34.48 & 60.57 & 43.95 & 35.27 & 61.37 & 44.79 \\
&Erased by Our & \textbf{1.14} & \textbf{23.45} & \textbf{2.17} & \textbf{1.21} & \textbf{23.45} & \textbf{2.29} & \textbf{12.58} & \textbf{70.00} & \textbf{21.33} & \textbf{13.46} & \textbf{70.62} & \textbf{22.61} \\
\hline
\multirow{4}{*}{EAST \shortcite{Zhou2017EAST}} & Original image & 51.22 & 79.10 & 62.18 & 52.50 & 79.14 & 63.12 & 70.10 & 81.50  & 75.37 & 70.70 & 81.61 & 75.77 \\
\cline{2-14}
&Erased by Pix2Pix \shortcite{isola2017image} & 21.03 & 77.12 & 33.05 & 21.78 & 77.44 & 34.00 & 10.19 & 69.45 & 17.78 & 10.37 & 69.45 & 18.05 \\
&Erased by Scene text eraser \shortcite{Nakamura2017Scene} & 24.30 & 40.30 & 30.32 & 24.74 & 40.61 & 30.75 & 10.08 & 39.09 & 16.03 & 10.35 & 39.09 & 16.37 \\
&Erased by Our & \textbf{0} & \textbf{0} & \textbf{0} & \textbf{0} & \textbf{0} & \textbf{0} & \textbf{5.66} & \textbf{73.42} & \textbf{10.51} & \textbf{5.75} & \textbf{73.42} & \textbf{10.67} \\
\hline
\end{tabular}
\end{table*}
\subsection{4.3. Comparison with state-of-the-art methods}
We compared the performance of the proposed EnsNet with the relevant and recent state-of-the-art methods: {\bf {Pix2Pix}} \cite{isola2017image} and {\bf {Scene text eraser}} \cite{Nakamura2017Scene}.

The results on the synthetic and real dataset are shown in Tab. \ref{tab:components}, and demonstrate that the proposed EnsNet can significantly outperform previous state-of-the-art methods \cite{isola2017image,Nakamura2017Scene} in all metrics.
Tab. \ref{tab:compareSOTA} adopts two different detection methods \cite{Zhou2017EAST} \cite{Fu2017DSSD} and two protocols: the DetEval and the ICDAR 2013 evaluation \cite{Karatzas2013ICDAR,Wolf2006Object} to further demonstrate the robustness of our method. The speeds of different methods are also listed in Tab \ref{tab:components}. Our method can achieve 333 fps on a i5-8600 CPU device. Although EnsNet is slightly slower than Pix2Pix \cite{isola2017image} , it is twice faster than the scene text eraser \cite{Nakamura2017Scene}.

\begin{figure}[htb]
\includegraphics[width=\linewidth,,height=10.0cm]{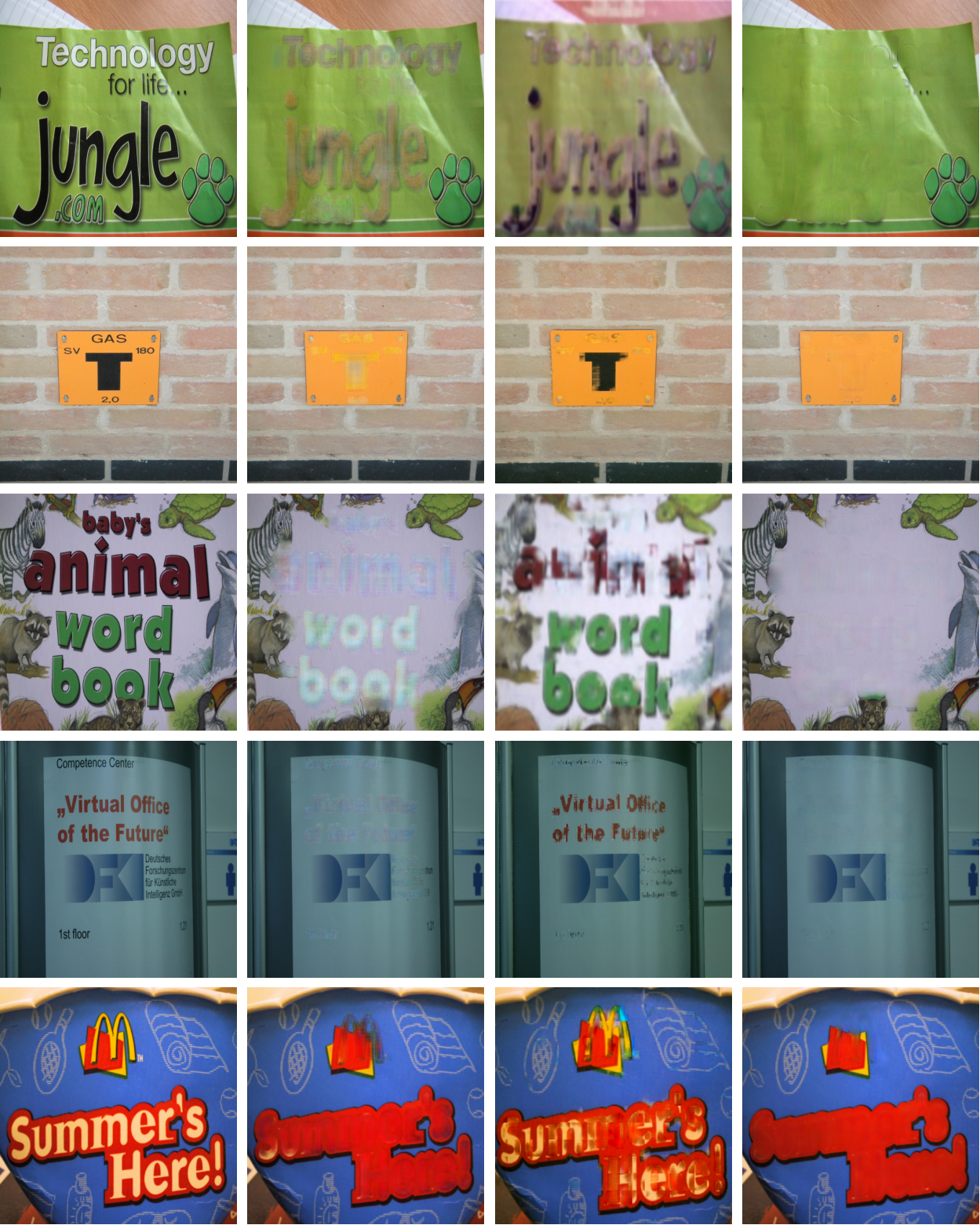}
\caption{Results of comparing a few different methods on sample images from a real dataset. Images from left to right: input image, Pix2Pix \cite{isola2017image}, Scene text eraser \cite{Nakamura2017Scene}, and our method.}
\label{fig:resCompareSOTA}
\vspace{0cm}
\end{figure}

Furthermore, we qualitatively compared the results generated by different methods, as shown in Fig. \ref{fig:resCompareSOTA}. Based on the observation, we found that Pix2Pix \cite{isola2017image} can only remove parts of the text region, and the restored regions contain many fish scale artifacts. As for the scene text eraser \cite{Nakamura2017Scene}, most of the text in the images can not be erased completely; meanwhile, some non-text backgrounds are also influenced. In contrast, the effectiveness of our method can be intuitively visualized in Fig. \ref{fig:resCompareSOTA}. The results demonstrate that the EnsNet can not only almost perfectly erase the text regions (even large-size text) but also maintain the reality of the backgrounds.

\subsection{4.4. Generalization to general object removal}
\begin{figure}[htb]
\includegraphics[width=\linewidth,height=4.0cm]{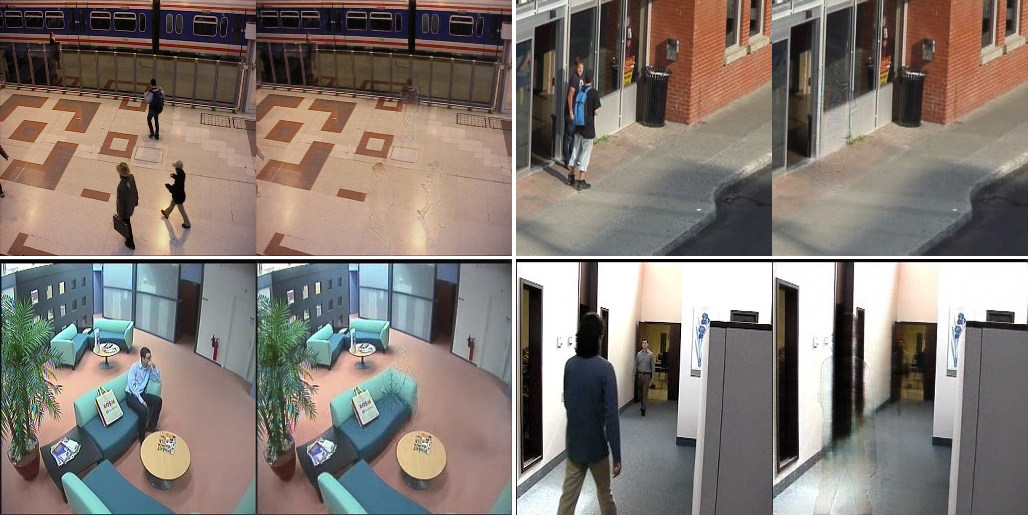}
\caption{ Experimental results on pedestrian removal. Images from left to right: input images and removal results.}
\label{fig:resPedestrian}
\end{figure}

In this section, an experiment on the general object removal task is conducted to test the generalization ability of EnsNet.
We evaluate EnsNet on a well-known Scene Background Modeling (SBMnet) dataset \cite{Jodoin2017Extensive}, which is targeted at background estimation algorithms.
However, SBMnet is a video-based dataset, implying that the methods can take advantage of the sequence frames to reconstruct all the background information. Therefore, it is not fair to quantitatively compare our method with other methods because our method uses only one static image as the input without any context information of other frames. It is noteworthy that this is a novelty of EnsNet that enables a plausible background to be reconstructed. The qualitative results are shown in Fig. \ref{fig:resPedestrian}. Without specific tuning, our method can still generalize excellently to the pedestrian removal task.
\section{5. CONCLUSION}
This paper proposes EnsNet - a novel method that can effectively erase text from scene text images.

All components of EnsNet are proved crucial to achieve good erasing results, including 1) lateral connection, which can effectively integrate high-level semantics and low-level detailed information; 2) a novel refined loss, which can remarkably reduce the artifacts; and 3) a simple but highly effective local-aware discriminator, which ensures the local consistency of the text-erased regions. We also constructed a new synthetic dataset for the scene text erasing task, which will be published for future research.

To the best of our knowledge, EnsNet is the first method that can be trained end-to-end to directly erase the text on a whole image level, thus achieving outstanding erasing performance on a synthetic dataset and the ICDAR 2013 dataset. It also significantly outperformed previous state-of-the-art methods in terms of all metrics.

Furthermore, a qualitative experiment conducted on SBMNet proved that our method could perform well in general object removal tasks, and further investigations would be carried out in future research.
\section{Acknowledgement}
This research is supported in part by GD-NSF (no. 2017A030312006), the National Key Research and Development Program of China (No. 2016YFB1001405), NSFC (Grant No.: 61472144, 61673182, 61771199), and GDSTP (Grant No.: 2017A010101027), GZSTP(no. 201607010227).
\bibliographystyle{aaai}
\bibliography{my_paper}
\end{document}